\documentclass{article}
\usepackage{spconf,amsmath,graphicx,hyperref}
\makeatletter
\let\oldthebibliography\thebibliography
\def\thebibliography#1{%
  \oldthebibliography{#1}%
  \fontsize{9pt}{10pt}\selectfont    
  \setlength{\itemsep}{0pt plus 0.2pt}%
  \setlength{\parskip}{0pt}%
  \setlength{\parsep}{0pt}%
}
\makeatother
\usepackage{pgfplots}

\usepackage{booktabs}
\usepackage{graphicx}
\usepackage{adjustbox}  
\pgfplotsset{compat=1.18}
\usepgfplotslibrary{groupplots}
\usepackage{tikz}
\usepgfplotslibrary{statistics}   
\usepackage{CJKutf8}

\newcommand{\smallheading}[1]{\vspace{0.1cm}\noindent\textbf{#1}\ \ }

\title{Adapting Language Balance in Code-Switching Speech}
\name{Enes Yavuz Ugan$^{\star }$ \qquad Ngoc-Quan Pham$^{\dagger}$ \qquad Alexander Waibel$^{\star\dagger}$}
\address{$^{\star}$Interactive Systems Lab, Karlsruhe Institut of Technology (KIT), Germany \\
      $^{\dagger}$InterACT, Carnegie Mellon University (CMU), USA}

\begin{document}
%
\maketitle
\begin{abstract}
Despite achieving impressive results on standard benchmarks, large foundational models still struggle against code-switching test cases. 
When data scarcity cannot be used as the usual justification for poor performance, the reason may lie in the infrequent occurrence of code-switched moments, where the embedding of the second language appears subtly.
Instead of expecting the models to learn this infrequency on their own, it might be beneficial to provide the training process with labels.

Evaluating model performance on code-switching data requires careful localization of code-switching points where recognition errors are most consequential, so that the analysis emphasizes mistakes occurring at those moments.
Building on this observation, we leverage the difference between the embedded and the main language to highlight those code-switching points and thereby emphasize learning at those locations.

This simple yet effective differentiable surrogate mitigates context bias during generation—the central challenge in code-switching—thereby improving the model’s robustness. Our experiments with Arabic and Chinese-English showed that the models are able to predict the switching places more correctly, reflected by the reduced substitution error.
\end{abstract}
\vspace{-0.5em}
\section{Introduction}
Code-switching, the alternation of two or more languages within a single utterance is a  common phenomenon in multilingual speech.
Since code-switching carries socio-linguistic significance and conveys additional meaning \cite{dougruoz2023survey, myers1977bilingual}, improving the processing of such data is imperative.
Accordingly, the topic has received growing attention across natural language processing (NLP), such as machine translation (MT) \cite{ahmed2018codes} and automatic speech recognition (ASR) \cite{liu2024aligning}.
Multilingual and cross-language modeling has long been studied as a way to leverage shared resources across languages \cite{schultz2001experiments,stuker2003integrating,stuker2003multilingual}.
More specifically, code-switching in ASR has gained attention for language pairs including Arabic-English \cite{hamed2020arzen, abdallah2024leveraging} (Egyptian- and Tunisian-English), Mandarin-English \cite{lyu2010seame,lovenia2021ascend}, Spanish-English \cite{weller2022end, deuchar2014building} and German-English \cite{ugan2024decm,ugan2025weight}.

With the intra-sentential code-switched data, embedded tokens are vastly outnumbered by those of the matrix language. The training loss is dominated by matrix-language errors, while there is not enough statistics to learn from the subtle switching points limiting improvements at the actual switch points. From the evaluation point of view, these code-switched elements are made insignificant when looking at the overall word error rate (WER). The Point-of-Interest Error Rate (PIER) metric designed for code-switching \cite{ugan2025pier} showed that they should be weighted higher in evaluation, and ideally also in training.

With that observation, the research question in our work is:
\emph{Can we design a differentiable surrogate loss aligned with PIER to reduce embedded language errors in code-switching ASR?}
Our key contributions are:
\begin{itemize}\setlength\itemsep{0.1em}
    \item We propose a token-weighted cross-entropy loss that emphasizes embedded tokens as a differentiable surrogate aligned with PIER. By utilizing the difference between the embedded and main language, we can guide the model to emphasize on the switching places. 
    \item Through empirical evaluations across diverse code-switching corpora, we found that the optimization with points of interest awareness can help the model detect and switch languages more accurately. 
\end{itemize}

\vspace{-0.5em}
\section{Related Work}
\smallheading{Point-of-Interest Error Rate (PIER)}
\label{rel:pier}
PIER refines the standard WER by measuring errors solely on designated “points of interest”, i.e. code-switched words ($\mathcal{I}$), thus focusing evaluation on the most challenging segments \cite{ugan2025pier}.  
PIER is defined as:
\begin{equation}\label{eq:pier}
\mathrm{PIER} \;=\; 
\frac{S_{\mathrm{I}} \;+\; D_{\mathrm{I}} \;+\; I_{\mathrm{I}}}
{N_{\mathrm{I}}}=\frac{|\mathcal{A}_I|}{N_{I}}\,,
\end{equation}
where \(S_{\mathrm{I}}\), \(D_{\mathrm{I}}\), and \(I_{\mathrm{I}}\) are the substitutions, deletions, and insertions at the points of interest, and \(N_{\mathrm{I}}\) is the total number of reference code-switched words.  
These errors are calculated as follows:
\begin{equation}\label{eq:a_i_algo}
\mathcal{A}_{\mathcal{I}} = 
\begin{cases} 
\mathcal{A}_{\mathcal{I}}^{*} \cup \left\{ o \in \mathcal{A} \mid i_\text{src}^o > |\text{REF}| \right\}, &\max(\mathcal{I}) = |\text{REF}| \\
\mathcal{A}_{\mathcal{I}}^{*}, & \text{otherwise}
\end{cases}
\end{equation}
with
\[
\mathcal{A}_{\mathcal{I}}^{*} = \left\{ o \in \mathcal{A} \mid i_\text{src}^o \in \mathcal{I} \right\}
\]
where $\mathcal{A}$ is the set of edit operations $o = (\;t\;, i_\text{src}\;, i_\text{res}\;)$ defining insertions, deletions and substitutions to transfer a hypothesis string into the reference string.
Thus, a subset of edit operations considered for calculating WER is used to obtain a more meaningful evaluation of under-represented code-switched words.


\smallheading{Modeling Code-Switching}
A variety of text-only augmentation strategies have been proposed to enrich code-switching language models. \cite{hussein2023textual} and \cite{hamed2022investigating} both explore lexical swaps into monolingual corpora, either randomly or driven by linguistic theory.
While \cite{hussein2023textual} relies on external translation and alignment tools to insert English words into Arabic text, it is able to reduce WER.
\cite{hamed2022investigating} finds that segment-based replacements maximize WER gains and that model-driven substitutions produce the most natural switches under human evaluation.
Other works extend augmentation into the acoustic domain. 
\cite{du2021data, farooq2025sage} splice existing code-switched utterances and synthesize speech after word-level translations or insertions, improving both language and acoustic models, though depending on parallel data and synthesis quality.
\cite{ugan2022language} show that simple inter-sentential concatenation of audio and text pairs boosts language agnostic ASR performance on both mixed language and monolingual test sets.
In \cite{nakayama2021code}, the authors introduce a speech-chain loop in which ASR and Text-to-Speech (TTS) models iteratively train each other, further reducing WER but at the cost of increased training complexity.

\smallheading{Handling Data Imbalance}
In \cite{henning2022survey} the authors analyze different cases of class imbalance and approaches for tackling them.
One of the approaches, sampling more data from the under-represented distribution, is not an option in our scenario.
Another approach based on data augmentation has been explored in previous work \cite{ugan2022language, du2021data,nakayama2021code}.
A further line of research in class imbalance is cost-sensitive learning, where misclassifications are penalized differently depending on a predefined cost structure \cite{1549828}.

Our work constitutes a special case of this paradigm: 
we adopt a weighted token-level cross-entropy loss, directly emphasizing under-represented language tokens.
This differs from earlier work jointly modeling ASR and language identification (LID) \cite{schultz1996lvcsr}, foreshadowing later multi-task approaches that incorporate LID as an auxiliary objective \cite{tseng2021mandarin,hou2020large}, which introduce additional heads or supervision.
In contrast, our method integrates language awareness into the primary ASR loss without requiring extra modules or parallel annotations.
\vspace{-0.5em}
\section{Problem Formulation}


In the context of code-switching ASR, where utterances alternate between two or more languages, treating every token identically during training fails to account for distributional and acoustic ambiguities at language boundaries.
We therefore introduce a script-aware loss weighting mechanism that systematically reweights gradient updates toward under-represented or harder to model tokens. 

\subsection{Point-of-Interest guided training}
Let $\{(x_i, y_i)\}_{i=1}^N$ denote the training pairs, where $x_i = (x_{i,1}, \dots, x_{i,L})$ is a sequence of $L$ speech features, and $y_i = (y_{i,1}, \dots, y_{i,T})$ is the corresponding token sequence of length $T$.
Let 
\[
s_{i,t} \in \{0,1\}
\]
be the token level script label: (0 = Matrix Language (ML), 1 = Embedded Language (EL)\footnote{One can also consider adding a mixed script label for the case of multiple scripts present in a single token.}. 
Standard cross‐entropy minimizes:
\begin{equation}
\label{eq:standardCELoss}
\mathcal{L}
= -\frac{1}{N T} \sum_{i=1}^N \sum_{t=1}^T \log p(y_{i,t} \mid x_i)\,,
\end{equation}
which assumes each position contributes equally. 
Instead, we define a weight function $w(s)$ with
\[
w(1) = \alpha > 1,\quad w(0) = 1,
\]
where $\alpha$ is a hyperparameter controlling the emphasis on embedded tokens, and reformulate the loss as:
\begin{equation}
\mathcal{L}_{\mathrm{weighted}}
= -\frac{1}{\sum_{i,t} w(s_{i,t})} \sum_{i=1}^N \sum_{t=1}^T w(s_{i,t}) \,\log p(y_{i,t} \mid x_i)\,.
\end{equation}
By choosing $\alpha > 1$, Latin‐script tokens, often less frequent in code‐switched corpora, receive a proportionally larger share of the gradient.
This drives the model to allocate representational capacity and learning focus on the most critical regions: the switching points.

\subsection{Integration into Training}
With that objective in mind, we focus on re-weighting the token-level labels, in order to help the model notice the code-switching positions during training.
In order to achieve the token-level weights, we construct a lookup table that maps each label ID to a script class.
Intuitively, we can differentiate code-switching tokens using their difference in written scripts.
To do this, we first decode all label IDs into their string representation and then apply regex-based rules to assign each token a script category.
In our experiments Section~\ref{sec:experiments} we focus on Arabic-English and Mandarin-English code-switching. 
Accordingly, the script categories are Latin (for English), Arabic, and Han. 
Other language pairs would require analogous script classification.
The results are stored in a tensor, mapping each token ID to its corresponding script class for efficient lookup.
During training we perform vectorized lookup:
batched decoder input IDs index the mapping tensor, yielding the $s_{i,t}$ mask aligned with target tokens.
For the weighted loss we first compute unreduced per-token losses $l_{i,t}$, mutliply them by $w_{s_{i,t}}$, and finally sum and normalize by $\sum(w)$.
Compared to complicated gating structures \cite{wang2023language, yang2024effective}, this substitution introduces negligible runtime overhead while injecting script sensitivity directly into the training objective.
\footnote{Another more general approach, we did not apply here, is a word level weighted loss instead of a token level.
This is particularly useful for language pairs with the same or similar writing systems, such as Vietnamese-English. Here, automatic language classification algorithms \cite{kargaran2024masklid} can be used.}

\vspace{-1.em}
\section{Experiments}
\label{sec:experiments}
In this section, we evaluate the effect of our proposed weighted token loss on different datasets and perform ablation studies for better analysis.
\vspace{-0.5em}
\subsection{Single Corpus}
\label{subsec:singlecorpus}
We first test our newly proposed loss on the ArzEn corpus \cite{hamed2020arzen}, which contains clean Egyptian-Arabic–English code-switched recordings.
In order to verify our hypothesis that weighting the code-switched elements differently can directly alter the learning process, we compare the following models.
\emph{Whisper} is the off-the-shelf \emph{large-v3-turbo} model; 
\emph{FT} is the same model fine-tuned with standard CE loss.
As a baseline using token-level labels, we apply multi-task learning with an auxiliary language classification head.
This model is trained using an additional language classification loss, weighted by $\alpha$, reported as MT$_{1.0}$ and MT$_{0.3}$ in Table~\ref{tab:singlecorpusarzen10}.
\begin{equation}
\mathcal{L}_{\text{total}} = \mathcal{L}_{\text{CE}} + \alpha \, \mathcal{L}_{\text{lang}}
\end{equation}
, with $\alpha=1$ and $0.3$ repsectively.
Our proposed model is denoted as $WL_{w(1)}$, where English tokens are treated as embedded language tokens.
We perform a hyperparameter search over multiple $w(1)$ values (1.3, 1.5, 1.7, 2).

\begin{table}[t]
\centering
\small
\begin{tabular}{crr}
  \toprule
  \textbf{$w(1)$} 
    & \textbf{WER (\%)} $\downarrow$  
    & \textbf{PIER (\%)} $\downarrow$  \\
    \midrule
    Whisper & 60.32    & 62.73   \\
    FT     & 31.37    & 19.55  \\
    MT$_{1}$ &    35.9    &   22.98   \\
    MT$_{0.3}$&    34.78   &   23.78   \\    
    \midrule
  WL$_{1.3}$         
    & 33.84 (+2.47) & 18.11 (–1.44)\\
  WL$_{1.5}$         
    & \textbf{31.19} (–0.18) & 18.29 (–1.26)\\
  WL$_{1.7}$         
    & 31.81 (+0.44) & \textbf{17.59} (–1.96)\\
  WL$_{2}$         
    & 38.21 (+6.84) & 22.00 (+2.45)\\
  \bottomrule
\end{tabular}%
\caption{ArzEn‐finetuning results. Unadapted Whisper (Whisper), standard CE fine-tuned variant (FT), weighted multi-task fine-tuned variants (MT), and our newly weighted CE variants (WL).}
\vspace{-0.8em}
\label{tab:singlecorpusarzen10}
\end{table}
It is noticeable that standard fine-tuning (FT) outperforms a multi-task fine-tuning approach (MT) for both $\alpha=1$ and $0.3$, on both metrics WER and PIER.
This suggests that simple multi-task learning does not necessarily help improving a strong foundational models code-switching performance.

Comparing the FT model with our weighted loss (WL) versions we can see constant improvements on the actual code-switched words for $w(1)=1.3,1.5,1.7$.
However, it is important to note that in most cases except for $w(1)=1.5$ we see a negative trade-off with general WER which indicates that our approach needs careful adjustments as a hyperparameter to achieve best results.
For the best weights $1.5$ and $1.7$ we get a relative improvement of 6.45\% and 10.03\% in terms of PIER score. The model outputs also have more reliable embedded (English) prediction than the fine-tuned model, so the stronger weights emphasized the correlation between the acoustic and surface tokens, to overpower the bias from the context, which is one of the main problems in code-switching recognition.  
However, the WER slightly degrades as the model encountered early termination for long utterances. 
We hypothesize that the model struggles to ``return'' to the main language.
Setting $w(1)=2$ yields significantly worse results on WER and PIER scores, looking at the hypothesis, we can see that the model visibly struggles to produce transcripts in monolingual Egyptian utterances. 
In longer utterances, in cases where the model misses a language switch, it also stops predicting some following words, thus increasing deletion errors.

With $w(1)=1.5$ or $w(1)=1.7$, the model achieves a near-Pareto optimal balance between WER and PIER, effectively trading off overall recognition accuracy and code-switching precision. 
\vspace{-0.5em}
\subsection{Ablation on different language pairs}
\label{subsec:ablation_seame}
Table~\ref{tab:corpusascend10} shows results on Mandarin-English code-switching, evaluated on both the in-domain ASCEND and out-of-domain SEAME test sets. 
\begin{table}
\centering
\small
\begin{tabular}{crrr}
  \toprule
  \textbf{$w(1)$} 
    & \textbf{Eval}
    & \textbf{MER ($\%$)} $\downarrow  $  
    & \textbf{PIER ($\%$)} $\downarrow$ \\
    \midrule
    Whisper &Ascend & 20.4   &  34.27    \\
            & Seame & 39.72 & 58.67 \\
    FT 
    &Ascend & 58.66 (13.32)   & 22.54    \\
    & Seame & 32.27 &   41.12 \\
    \midrule
  WL$_{1.3}$   &Ascend & 66.67 (12.85)  & 21.06 \\
        &Seame  & 31.82 & 39.88 \\
  WL$_{1.5}$   &Ascend & \textbf{15.6} (13.44) & 21.91  \\
        &Seame & 29.26  & 41 \\
  WL$_{1.7}$   &Ascend & 70.13 (\textbf{12.79})   & 20.95    \\
        &Seame  & 33.43 &   \textbf{39.62} \\
  WL$_{2}$   & Ascend & 32.37 (16.58)    &  \textbf{20.53}    \\
        & Seame & 30.92 & 41.01 \\
  \bottomrule
\end{tabular}%
\caption{ASCEND‐finetuning results in Mixed Error Rate (MER) and PIER. Unadapted Whisper (Whisper), standard CE fine-tuned (FT), our proposed weighted versions (WL). Outside brackets the MER is reported (hallucination-free MER reported in brackets). We report cross-corpus results on SEAME test data.}
\vspace{-0.5em}
\label{tab:corpusascend10}
\end{table}
We observed that fine-tuned models tend to produce extreme hallucinations in very short utterances. Accordingly, we report both overall Mixed-Error-Rate (MER) and hallucination-free MER (in brackets), computed by excluding hypotheses longer than 10$\times$ the reference length. Because hallucinations occurred only for single-character utterances, PIER is unaffected.
For example instead of \begin{CJK*}{UTF8}{gbsn}{嗯 (um)}\end{CJK*} the model produced a long nonsensical sequence \begin{CJK*}{UTF8}{gbsn}{一直到较适合适合合适业的选择合选...} \footnote{full hypothesis: 一直到较适合适合合适业的选择合选合选拼合选选拥合选用选拨选拘运运选拴运逐选运拨逐运权选拐材选逐权权逐杆逐材权用选权材逐杈权逼权杆选材杆权逆权杀权递权杉权逮权}\end{CJK*}.
When fine-tuned with standard CE loss (FT), the model's performance on ASCEND drastically degrades in terms of overall MER (58.7\%) due to hallucinations. 
However, the hallucination-free MER is 13.3\% much lower then Whisper.
This indicates that, aside from a few problematic utterances, the model adapts well to ASCEND. On SEAME, FT gives MER of 32.2\%, showing clear improvements over the off-the-shelf model along with a reduced PIER (41.1\% vs. 58.67\%). 
Thus fine-tuning helps cross-corpus transfer, but hallucinations remain a critical issue.
The WL variants generally maintain the strong in-domain gains of FT while in some cases also reducing hallucination issues and improving cross-corpus robustness.
With $w(1)=2$ we can see the best PIER score on ASCEND (20.53\%), however, looking at the hypothesis we see similar, as discussed in Section~\ref{subsec:singlecorpus}, the model struggles with monolingual mandarin utterances.
$w(1)=1.5$ achieves the lowest overall MER on ASCEND improving over the baseline by 23.53\% MER. 
On Seame, it gives the best cross-corpus MER reducing over Whisper and FT models by 26.33\% and 9.33\% MER. 
It also shows slightly improved code-switching capabilites with PIER 21.9\% on ASCEND.
The slightly increased weight $1.7$ produces best hallucination-free MER (12.8\%), and improved PIER compared to the FT model, but has high overall MER on ASCEND due to hallucinations.

Whisper provides a reasonable starting point but with high MER and PIER. 
Standard fine-tuning (FT) reduces PIER and improves hallucination-free MER, though overall MER is unstable. 
Our weighted approach demonstrates that the choice of weight is crucial: WL$_{1.5}$ not only lowers MER on both ASCEND and SEAME but also further reduces PIER on ASCEND compared to FT, making it the most balanced and effective configuration, while other weights yield mixed outcomes.

\vspace{-0.5em}
\subsection{PIER Breakdown}
\label{subsec:pierbreakdown}
\begin{figure}[t]
\centering
\begin{tikzpicture}
\begin{axis}[
    ybar,
    bar width=9pt,
    width=.9\textwidth,
    height=6.0cm,
    symbolic x coords={I,D,S},
    xtick=data,
    xtick style={draw=none},
    ylabel={Error Count},
    ymin=0,
    ymax=4000,
    nodes near coords,
    nodes near coords align={vertical},
    every node near coord/.append style={font=\scriptsize},
    xticklabel style={font=\normalsize},
    axis x line*=bottom,
    axis y line*=left,
    enlarge x limits=0.2,
    x=2.5cm,
    legend style={
        at={(0.5,-0.15)},
        anchor=north,
        legend columns=2,
        font=\small
    },
      legend image code/.code={
      \draw[#1] (0cm,-0.12cm) rectangle (0.28cm,0.12cm);
  },
]

\addplot+[ybar, fill=blue!40] coordinates {(I,86) (D,235) (S,437)};
\addplot+[ybar, fill=blue!80] coordinates {(I,82) (D,222) (S,405)};

\addplot+[ybar, fill=orange!40] coordinates {(I,307) (D,1203) (S,3444)};
\addplot+[ybar, fill=orange!80] coordinates {(I,368) (D,1209) (S,3347)};

\legend{
    Embedded (FT),
    Embedded (WL$_{1.5}$),
    Matrix (FT),
    Matrix (WL$_{1.5}$)
}
\end{axis}
\end{tikzpicture}
\caption{Breakdown of insertion (I), deletion (D), and substitution (S) error counts on Arzen. We compare the baseline fine-tuned model (FT) with our WL$_{1.5}$, separately for embedded and matrix words.}
\vspace{-0.5em}
\label{fig:arzen_pier_breakdown}
\end{figure}
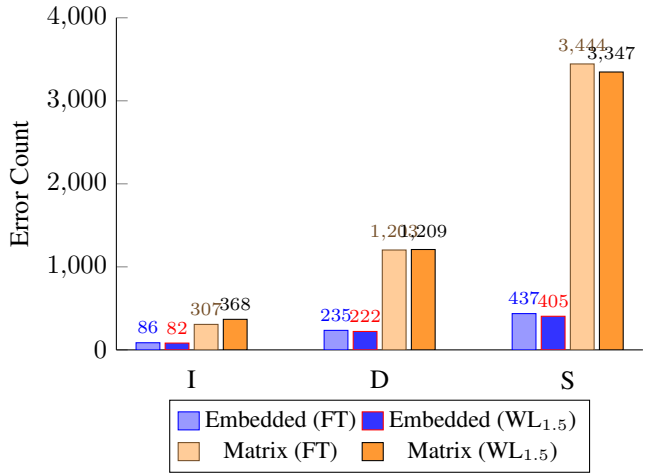

Figure~\ref{fig:arzen_pier_breakdown} decomposes PIER into insertion, deletion and substitution errors separately for embedded and matrix language words, comparing the baseline fine-tuned model (FT) with our weighted variant at $w(1)=1.5$, taken from our experiments in Section~\ref{subsec:singlecorpus}.

For embedded langauge words, we see consistent improvements: insertions decrease slightly (86$\rightarrow$82), deletions are reduced (235$\rightarrow$222), and subsitutions drop more noticably (437$\rightarrow$405). 
This confirms that weighting the embedded language tokens leads to more reliable recognition of code-switched tokens, particularly reducing substitution errors, which are dominant.
For matrix words, the picture is more mixed. 
Substitutions are reduced from 3444 to 3347, showing that weighting does not harm recognition of the dominant language and may even help. 
Deletions remain essentially unchanged (1203$\rightarrow$1209), while insertions increase moderatly (307$\rightarrow$368). 
This suggests that improvements on embedded language tokens come at a small cost of additional insertions in the matrix language, but without degrading substitution accuracy.

Overall, the breakdown supports our main finding: $WL_{1.5}$ improves code-switching performance by reducing substitution and deletion errors on embedded words, while maintaining stable accuracy on the matrix language.
\vspace{-0.5em}
\section{Conclusion} 
In this work we proposed a simple but effective extension to cross-entropy fine-tuning by introducing token-level weighting targeted at embedded language tokens.
Across ArzEn and ASCEND/SEAME, WL$_{1.5}$ reduces PIER while keeping WER/MER stable, whereas WL$_{1.7}$ trades a small WER/MER increase for further PIER gains.
The error breakdowns show fewer substitution and deletion errors in embedded language words, with the accuracy in the matrix language largely preserved.
The method is lightweight, generalizes across corpora with no inference overhead.
Future work will explore broader language pairs and adaptive strategies for setting the weighting parameter.
%

\section{Acknowledgment}
\vspace{-0.5em}
This work was supported in part by BMBF (01EF1803B - RELATER, HoreKa), the EU Horizon program (101135798 – Meetween; 101213369 – DVPS), and grants from Zoom VC (2nd author) and Interactive-AI.
\vspace{-1em}


\bibliographystyle{IEEEbib}
\bibliography{strings,refs}

\end{document}